\documentclass[letterpaper, 10pt, conference]{IEEEtran}
\IEEEoverridecommandlockouts	

\usepackage{multirow}
\usepackage{lipsum}
\usepackage{amsmath}
\usepackage{amssymb}
\usepackage{subcaption}
\usepackage{graphicx}
\graphicspath{{./figures/}}
\usepackage{glossaries}
\usepackage[table,xcdraw]{xcolor}
\usepackage[ruled,vlined]{algorithm2e}
\usepackage[final]{listings}
\usepackage{longtable}
\usepackage{cite}

\usepackage{tikz}
\usepackage{textcomp}
\usepackage{hyperref}
\usepackage{lipsum}

\newcommand\copyrighttext{%
  \footnotesize \textcopyright 2022 IEEE.  Personal use of this material is permitted. Permission from IEEE must be obtained for all other uses, in any current or future media, including reprinting/republishing this material for advertising or promotional purposes, creating new collective works, for resale or redistribution to servers or lists, or reuse of any copyrighted component of this work in other works.
  DOI: 110.1109/ICVES56941.2022.9986939}
\newcommand\copyrightnotice{%
\begin{tikzpicture}[remember picture,overlay]
\node[anchor=south,yshift=10pt] at (current page.south) {\fbox{\parbox{\dimexpr\textwidth-\fboxsep-\fboxrule\relax}{\copyrighttext}}};
\end{tikzpicture}%
}

\usepackage[normalem]{ulem}
\usepackage{tikz,xcolor,hyperref}
\definecolor{lime}{HTML}{A6CE39}
\DeclareRobustCommand{\orcidicon}{
	\begin{tikzpicture}
	\draw[lime, fill=lime] (0,0) 
	circle [radius=0.16] 
	node[white] {{\fontfamily{qag}\selectfont \tiny ID}};
	\draw[white, fill=white] (-0.0625,0.095) 
	circle [radius=0.007];
	\end{tikzpicture}
	\hspace{-2mm}
}
\foreach \x in {A, ..., Z}{\expandafter\xdef\csname orcid\x\endcsname{\noexpand\href{https://orcid.org/\csname orcidauthor\x\endcsname}
			{\noexpand\orcidicon}}
}

\definecolor{mygray}{rgb}{0.5,0.5,0.4}
\definecolor{mygreen}{rgb}{0.2,0.5,0.2}
\definecolor{myblue}{rgb}{0.2,0.2,0.9}
\definecolor{mykwdclr}{rgb}{0.2,0.6,0.6}

\lstdefinestyle{mybase} {
	language=[Sharp]C,
	breaklines=true,
	showstringspaces=false,
	basicstyle=\small\tt,
	frame=single,
    numbers=left,
    numbersep=8pt,
	numberstyle=\tiny\color{mykwdclr},
	keywordstyle=\color{myblue},
	keywords=[2]{Mathf,Laser_Scanner,MonoBehaviour,Vector3,RaycastHit,Debug,Color,Physics}, 
	keywordstyle=[2]\color{mykwdclr}, 
	commentstyle=\color{mygreen}\ttfamily
}

\lstdefinestyle{mycsh} {
	style=mybase
}

\newacronym{ADAS}{ADAS}{Advanced Driving Assistant Systems}
\newacronym{LKA}{LKA}{Lane Keeping Assistance}
\newacronym{LDW}{LDW}{Lane Departure Warning}
\newacronym{LCA}{LCA}{Lane Change Assistant}
\newacronym{ADS}{ADS}{Automated Driving Systems}
\newacronym{CNN}{CNN}{Convolutional Neural Network }
\newacronym{ROS}{ROS}{Robot Operating System}
\newacronym{PCL}{PCL}{Point Cloud Library}
\newacronym{SAE}{SAE}{Society of Automotive Engineers}
\newacronym{DDT}{DDT}{Dynamic Driving Task}
\newacronym{ODD}{ODD}{Operational Design Domain}
\newacronym{LDS}{LDS}{Laser Distance Sensor}
\newacronym{LIDAR}{LIDAR}{Light Detection And Ranging}
\newacronym{RANSAC}{RANSAC}{Random sample consensus}
\newacronym{FIR}{FIR}{Finite Impulse Response}  
\title{\LARGE \bf Road Markings Segmentation from LIDAR Point Clouds using Reflectivity Information}

\author{Novel Certad\orcidN{} \emph{Graduate Student Member, IEEE}, Walter Morales-Alvarez\orcidW{} \emph{Student Member, IEEE}, \\and Cristina Olaverri-Monreal\orcidC{} \emph{Senior Member, IEEE}%
\thanks{Chair Sustainable Transport Logistics 4.0, Johannes Kepler University Linz, Altenberger Straße 69, 4040 Linz, Austria.
\texttt{\{novel.certad\_hernandez, walter.morales\_alvarez, cristina.olaverri-monreal\}@jku.at}}%
}

\begin{document}

\maketitle

\copyrightnotice
\thispagestyle{empty}
\pagestyle{empty}
\captionsetup[figure]{name={Fig.},labelsep=period}
\begin{abstract}
    Lane detection algorithms are crucial for the development of autonomous vehicles technologies. The more extended approach is to use cameras as sensors. However, LIDAR sensors can cope with weather and light conditions that cameras can not. In this paper, we introduce a method to extract road markings from the reflectivity data of a 64-layers LIDAR sensor. First, a plane segmentation method along with region grow clustering was used to extract the road plane. Then we applied an adaptive thresholding based on Otsu's method and finally, we fitted line models to filter out the remaining outliers. The algorithm was tested on a test track at 60km/h and a highway at 100km/h. Results showed the algorithm was reliable and precise. There was a clear improvement when using reflectivity data in comparison to the use of the raw intensity data both of them provided by the LIDAR sensor. 
\end{abstract}

\section{Introduction}
\label{sec:introduction}

Nowadays, lane-detection algorithms are crucial for the implementation of \gls{ADAS} with different levels of autonomy such as \gls{LKA}, \gls{LCA}, and \gls{LDW} among others. The development and deployment of vehicles with level 3 automation or higher (according to the automation levels represented in the Society of Automotive Engineers (SAE) J3016 standard \cite{sae}, ranging from ``no driving automation" (level 0) to ``full driving automation" (level 5)) makes them even more important.

The estimation of the lane's shape in structured roads very often relies on the white lines used as road markers and sometimes in the road curb itself. To their detection, most of the related works tend to process images taken by cameras located on the outside of the vehicle or installed under the windshield \cite{waykole2021,Lipski2008,chae2018}. This image-processing approach exhibits great results under both, good lighting and weather conditions. However, it fails at nighttime, in bright sunlight, or under adverse weather conditions.

However, \gls{LIDAR} sensors remain almost unaffected by poor lighting conditions \cite{martirena2020} and provide an accurate and reliable way to measure the distances to objects around the vehicle. \gls{LIDAR} prices have been dropping over the last few years resulting in a wider deployment and use in the autonomous vehicles field.

\begin{figure}[t]
	\centering
	\begin{subfigure}{0.237\textwidth}
		\includegraphics[width=\textwidth]{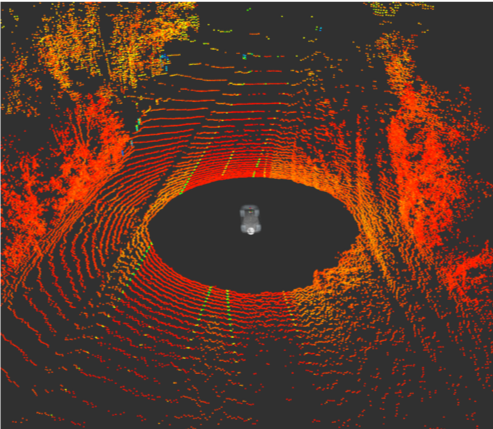}
		\caption{}
	\end{subfigure}
    \begin{subfigure}{0.237\textwidth}
		\includegraphics[width=\textwidth]{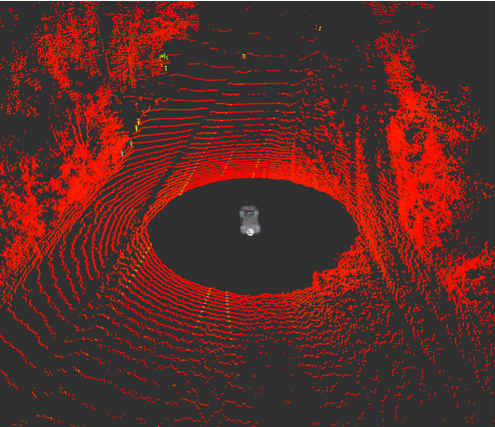}
		\caption{}
	\end{subfigure}
	\caption{The same point cloud from the test track was colored based on the level of reflectivity (a) and level of intensity (b).}
	\label{fig:channelcomparison}
\end{figure}

Since the lane-markings reflectivity is improved by reflective glass beads embedded in the surface of the paint, the intensity of the reflected laser beam is expected to be higher than the intensity of the beams reflected by the rest of the road (asphalt or concrete) \cite{martirena2020,lim2017,huang2021,ghallabi2018,Lindner2009}. However, the beam intensity is also affected by the distance to the target and the incidence angle against the surface. In recent years, \gls{LIDAR}'s manufacturers like Velodyne Lidar and Ouster have begun to offer reflectivity information along with the intensity and range signals. Reflectivity data indicates information about the inherent reflective property of the target, being not affected by lighting conditions and range \cite{thuy2010}. Therefore reflectivity is a powerful tool for road markings detection as can be seen in  Figure~\ref{fig:channelcomparison}.

In this paper, we review several road-marking detection algorithms (section~\ref{sec:relatedwork}) primarily based on \gls{LIDAR} intensity, range, or both. Later on, we propose our method to segment road-markings from \gls{LIDAR} point clouds using the reflectivity information instead of the intensity channel. The whole procedure was implemented using the \gls{PCL} \cite{pcl} in C++ ensuring compatibility with \gls{ROS} among other common frameworks. The detailed description is in section~\ref{sec:implementation}. The procedure was tested with two different datasets (section~\ref{sec:experiments}) and the results  (sections~\ref{sec:results}) demonstrate that the use of reflectivity information provides better results than intensity. Finally, the section~\ref{sec:conclusion} concludes the present work outlining future research.

\section{Related Work}
\label{sec:relatedwork}

Vision-based works are not included in this section as they are outside of the scope of the study. Nevertheless, a full review can be found in \cite{waykole2021}. 

In \cite{Lindner2009} a 6-layers \gls{LIDAR} sensor, with only three layers facing the ground, was used. A hybrid approach using range and intensity information was presented and the Otsu's method \cite{otsu1979} was used to distinguish between the road surface and lane marker signals \cite{otsu1979}. A detailed analysis regarding the material of the road (asphalt vs concrete), the type of road markings (plain vs raised), and different weather conditions (rain, sun, and night) were presented as results. However, the study did not present quantitative results.

In \cite{martirena2020} the authors presented a method for offline annotation of lane-markings. The lane-marking candidate points were detected as local maxima with an intensity value greater than a dynamic threshold. Odometry data was used to keep track of successive scans and the final candidates were then classified between solid or dashed lane-markings.

Both studies \cite{thuy2010} and \cite{li2013} achieved lane-markings detection using reflectivity information and tracking based on Extended Kalman filter. In \cite{thuy2010}, authors initially stacked consecutive one-dimensional \gls{LIDAR} scans to generate a reflectivity map or image, which was processed using common image processing techniques (binarization, Canny filtering, fixed thresholding). Afterward, they generated the underlying lane model by approximating the filtered points using the Hough transform. In \cite{li2013}, authors used a \gls{FIR} bandpass filter on a 64-layer \gls{LIDAR} sensor to remove noise. Then a fixed threshold was applied to obtain candidate line points. Finally, the authors fitted these points to clothoid curves to obtain the road markings.

In \cite{jung2018}, the authors detected the road-marking points using a fixed threshold over the intensity channel of a 32-layers \gls{LIDAR} sensor. First, the authors detected the road- lines by searching for a set of parallel lines integrated into a digital map using a GNSS/INS system. They then used an expectation-maximization method to detect parallel lines. The same \gls{LIDAR} was used in \cite{hata2014}, where a modified version of Otsu's method was applied to each scanned line. Then, a localization method based on the extracted road markings was presented. Finally, the resulting algorithm was improved in \cite{hata2014} and validated with a real test.
 
In \cite{huang2021} and \cite{huang2021b}, the intensity scans from an RS-LiDAR-16 were processed with a multi-threshold variation of Otsu's method to determine road-markings candidate points. These candidates were then filtered with a \gls{RANSAC} line model.

In \cite{ghallabi2018}, the authors proposed a method based on map localization. The method relied on road lane markings detection. Under the assumptions of flatness and smoothness of highway road surfaces, a ring analysis was performed to determine the points over the road surface. Then, an intensity-based thresholding was applied to extract the road marking points. Finally, these detected lane markings were matched to an HD map using a Particle Filter (PF). 


In \cite{uzer2019} the authors used \gls{LIDAR} scans to determine virtual lanes not related to the real lane-markings. First, a height-based filter eliminated the points of the road surface. Then, an unsupervised segmentation algorithm was run to cluster the points. Clusters with similar lateral positions were merged and two independent circular models were fitted to the clusters forming the so-called virtual lanes. A similar approach for unstructured roads was presented in \cite{han2014}.

The authors of \cite{lin2021} used a roadside \gls{LIDAR} for lane detection. The approach consisted of identifying the ground plane, extracting the lane marking points based on the difference in laser intensity, and dividing the ground within the range of \gls{LIDAR} scanning into different stripes to extract the lane markings.

\section{Implementation}
\label{sec:implementation}

\begin{figure}[t]
	\centering
	\includegraphics[width=0.3\textwidth]{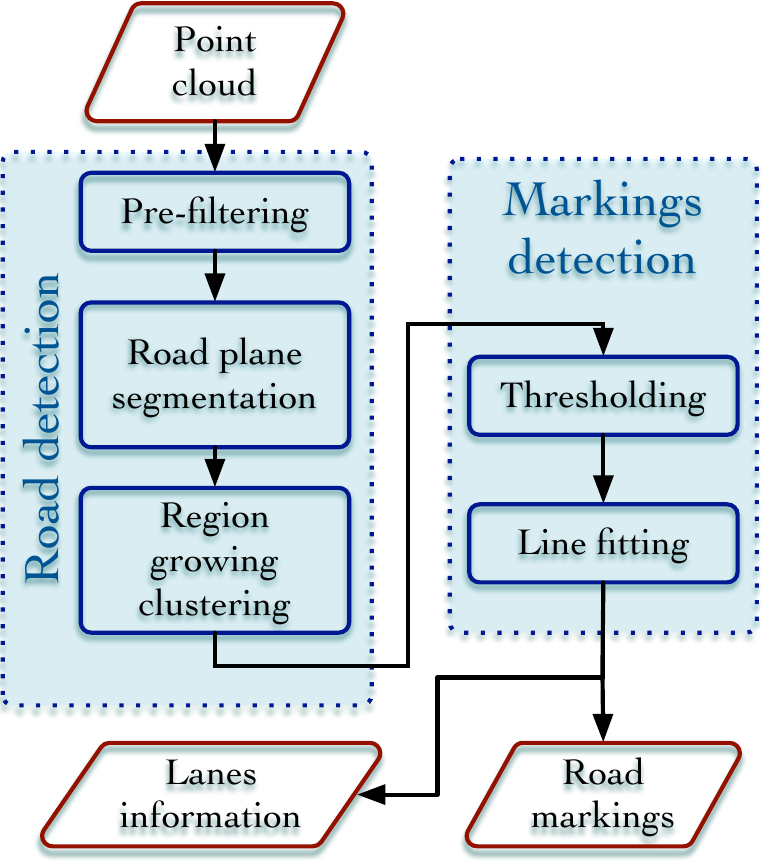}
	\caption{Flow graph of the developed algorithm.}
	\label{fig:diagram}
\end{figure}

Our developed procedure comprises two main blocks with several sub-processes. The complete procedure is depicted in Figure \ref{fig:diagram} and the detailed explanation is depicted in the sections below.

Consider a \gls{LIDAR} sensor with $N_L$ layers and $N_P$ points per layer. A single scan of the \gls{LIDAR} sensor produces a pointcloud $\mathbf{P^{A}}$ defined as a $N_L \times N_P$ set of points $\mathbf{p_{ij}} \in \mathbb{R}^6$:

\begin{equation}
    \mathbf{p_{ij}}= \left[ \begin{array}{cccccc}
x_{ij} & y_{ij} & z_{ij} & r_{ij} & I_{ij}  & R_{ij}  \end{array} \right]
\end{equation}

For each point $\mathbf{p_{ij}} \in \mathbf{P^{A}} $, the following quantities are defined:
  
\begin{itemize}
   
    \item  $x_{ij},y_{ij}, z_{ij}$ are the 3D spatial coordinates of the point from the \gls{LIDAR} reference frame.
    \item  $r_{ij}$ is the distance (range) from the \gls{LIDAR} reference frame to the point.
    \item  $I_{ij}$ and $R_{ij}$ are the intensity and reflectivity levels respectively.
    \item  $i$: is the $i$-th layer in which the point is located, indicated by an integer between  $1$ and $N_L$ 
    \item  $j$: is the $j$-th point index within the layer indicated by an integer between $1$ and $N_P$.
\end{itemize}

\subsection{Pre-filtering}
Point clouds obtained from multi-layer 3D-\gls{LIDAR} sensors contain massive amounts of data. Thus, it is important to reduce the size in order to improve the processing speed in further steps. Getting rid of non-relevant information is the best way to achieve this reduction. Therefore, we implemented a pre-filtering method that relied on the following two sources of spatial information for each point in the point cloud:

\begin{itemize}
    \item \textbf{Layers reduction:} We first filtered out the \gls{LIDAR} layers that were scanning above the horizon, as they acquired data from objects in the environment that are utterly irrelevant for the algorithm (buildings, trees, traffic signs, etc). In our implementation, we used only the 30 lower layers of our 64-layers \gls{LIDAR}.
    \item    \textbf{Height-based filter:} Similarly, we performed a filtering over the z-axis (perpendicular to the floor), by keeping the points between a lower ($z_{L}$) and upper threshold ($z_{U}$).
\end{itemize}

Consider $\mathbf{P^{A}}$ as the original pointcloud, then, the pre-filtered pointcloud $\mathbf{P^{B}}\subset \mathbf{P^{A}}$ is a set obtained by the following rule:

\begin{equation}
\begin{split}
   \mathbf{P^{B}} = \{ \mathbf{p_{ij}} \in \mathbf{P^{A}}:&  Z_{L} \leq z_{ij} \leq Z_{U}, 0 \leq i < 30,\\ 
   &0 \leq j < N_P \}
\end{split}
\end{equation}  

\subsection{Road plane segmentation}
Once the points are limited to the vicinity of the road plane, we used \gls{RANSAC} as an iterative method to find the plane model ($a,b,c,d$) which best fits the point cloud $\mathbf{P^{B}}$. We then filtered the points that did not fit into the plane model within a threshold ($TH_{plane}$) obtaining a new pointcloud $\mathbf{P^{C}}$ defined as:

\begin{equation}
\begin{split}
   \mathbf{P^{C}} = \{ \mathbf{p_{ij}} \in \mathbf{P^{B}} : &  
   |ax_{ij} + by_{ij} + cz_{ij} + d| \leq TH_{plane}\}
\end{split}
\end{equation} 

\begin{figure}[t]
	\centering
	\begin{subfigure}{0.237\textwidth}
		\framebox[\textwidth]{\includegraphics[width=\textwidth]{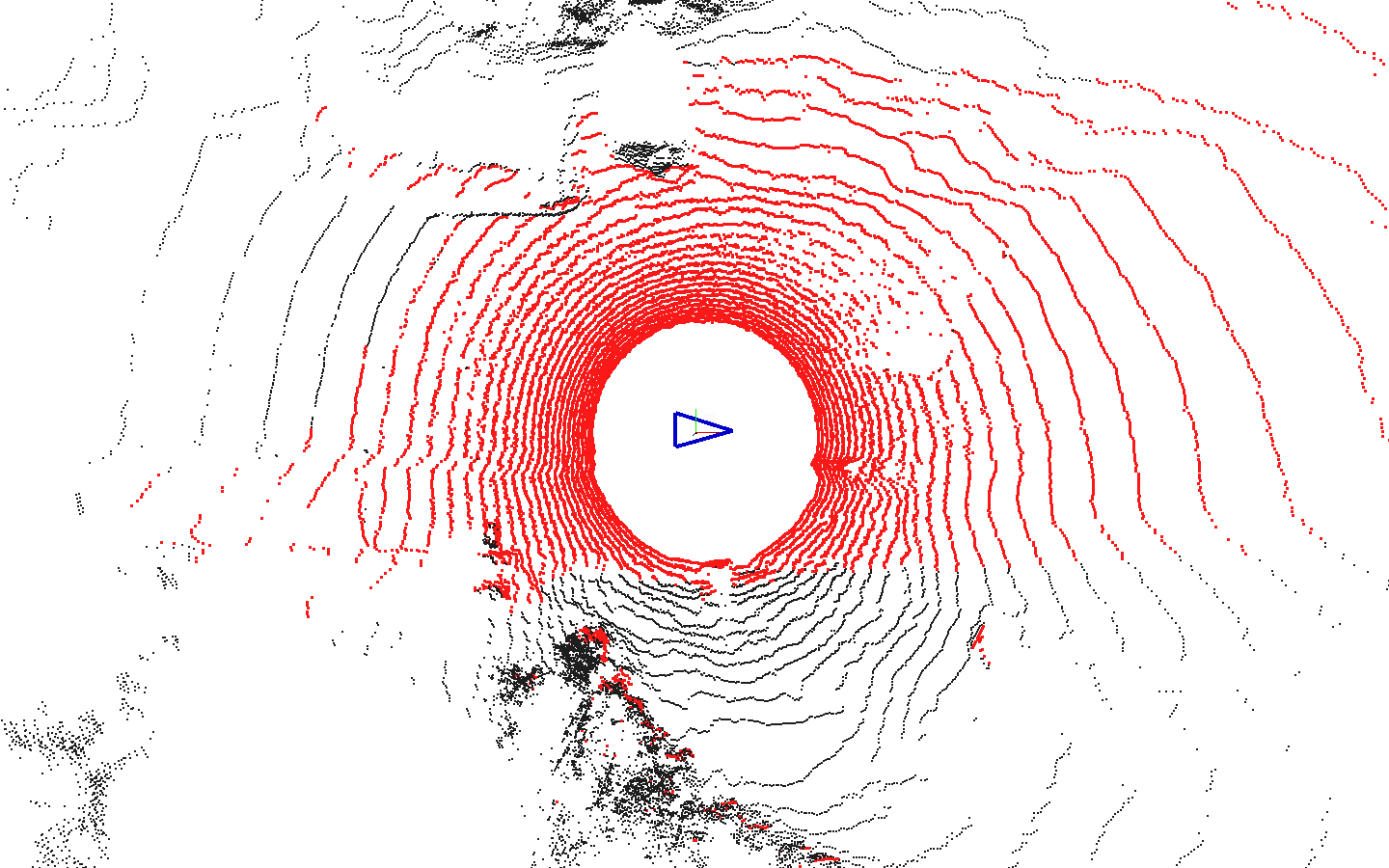}}
		\caption{}
	    \label{fig:alg_a}
	\end{subfigure}
	\begin{subfigure}{0.237\textwidth}
		\framebox[\textwidth]{\includegraphics[width=\textwidth]{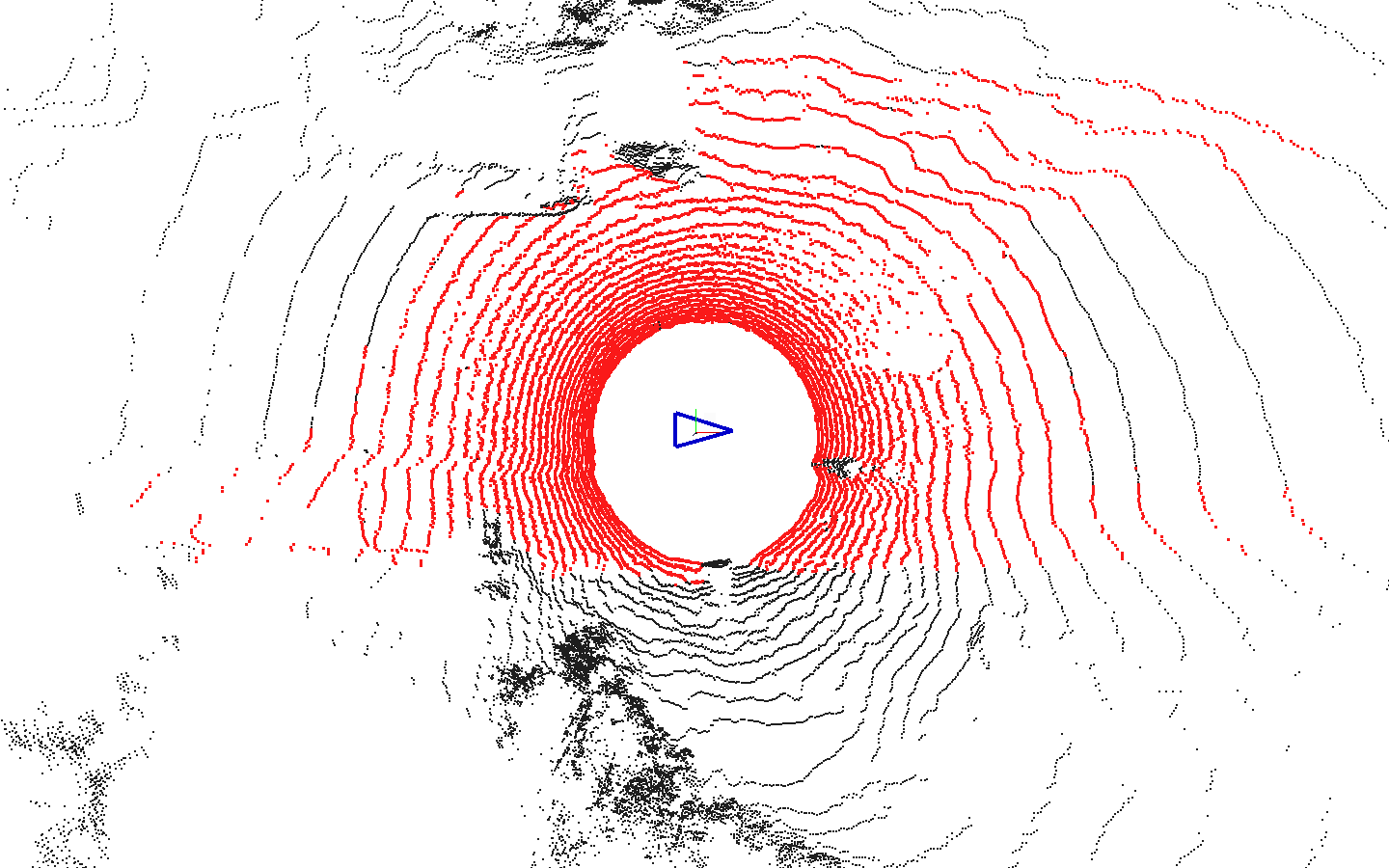}}
		\caption{}
	    \label{fig:alg_b}
	\end{subfigure}
	\begin{subfigure}{0.237\textwidth}
		\framebox[\textwidth]{\includegraphics[width=\textwidth]{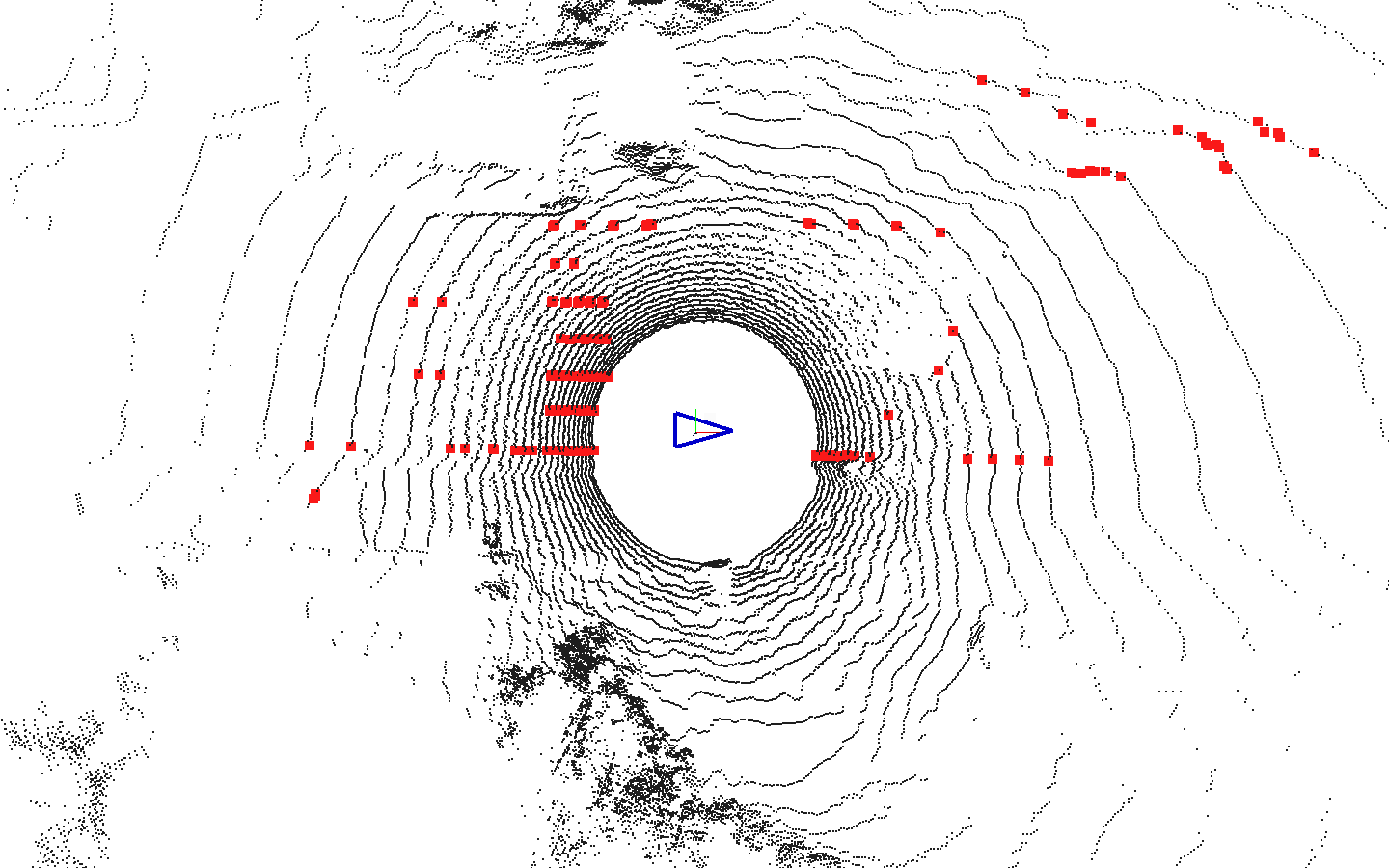}}
		\caption{}
	    \label{fig:alg_c}
	\end{subfigure}
    \begin{subfigure}{0.237\textwidth}
		\framebox[\textwidth]{\includegraphics[width=\textwidth]{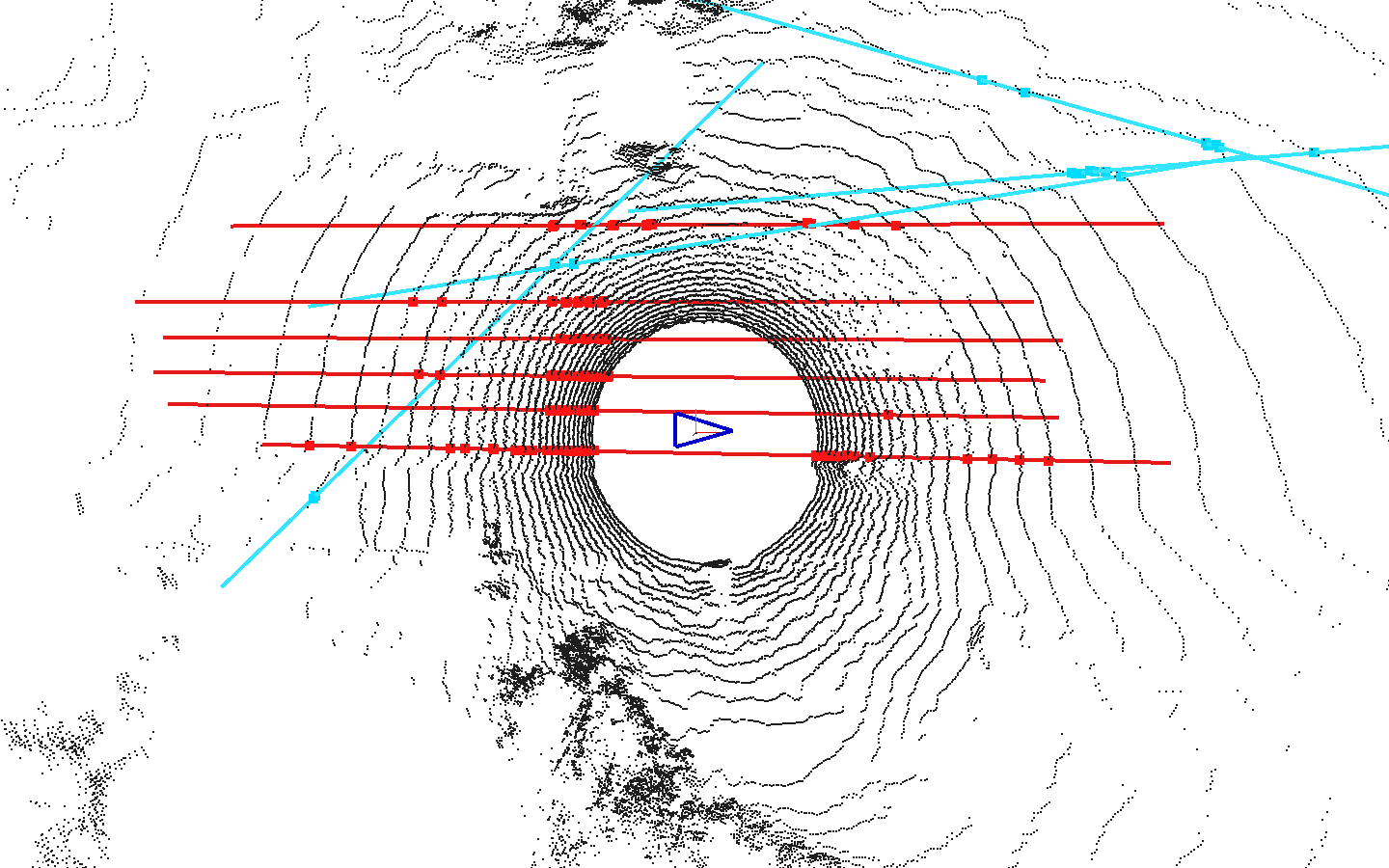}}
		\caption{}
	    \label{fig:alg_d}
	\end{subfigure}
	\caption{Point cloud output at the different stages of the developed algorithm. (a) In red is the output of the pre-filtering and plane segmentation stage. The output (red) of the region growing algorithm is depicted in (b). In (c) we can see the road-markings candidates (red) after the adaptive thresholding. Finally, in (d) we can see the final road marking (red points) with the associated lines (red lines) as well as the ones that were rejected by the algorithm (light blue). }
	\label{fig:algorithm}
\end{figure}

An example of the results after executing the pre-filtering along with the plane segmentation can be seen in Figure \ref{fig:alg_a}. 

\subsection{Region growing clustering}

A region-growing clustering was implemented in order to filter out points that did not belong to the road but the curb, the sidewalk, or other low-height structures. For each point, we analyzed the vicinity to find the $K_{RGC}$ nearest neighbors and then estimated the normal to the neighbors' surface evaluated in the original point. Then, the clustering algorithm selected a point in the cloud and started growing a region iteratively based on two thresholds. When the angle between two points' normals was less than $TH_{angle}$ and the difference between their curvatures was less than $TH_{curve}$, they were considered to be in the same region. An example of the resulting point cloud after the execution of the clustering is depicted in Figure \ref{fig:alg_b}.

\subsection{Adaptive thresholding}
\label{sec:thresholding}

Since the \gls{LIDAR} layers ($i$) could differ in calibration, the reflectance of the same object could produce different values in two different layers. Thus, a different threshold was calculated for each one of the layers ($i$). First, a reflectivity histogram ($h_{i}(k)$) of size $N_{bins}$ was built considering the $N_{Pi}$ points inside each one of the layers ($i$).

\begin{equation}
    h_{i}(k) = \left. \sum_{j=0}^{N_{Pi}}\dfrac{1}{N_{Pi}}\right|_{R_{ij}=k} k = 1...(N_{bins})
\end{equation}

At this point, a threshold was applied across the reflectivity channel to split the remaining data points between two possible classes: road ($C_R$) and road markings ($C_M$). As seen in the literature, Otsu's method \cite{otsu1979} is widely used to find the threshold that maximizes the inter-class variance ($\sigma_{b}^2(t)$) (or in other words minimizes the intra-class variance) \cite{huang2021,Lindner2009,hata2014,huang2021b}.
\begin{equation}
\sigma_{b}^2(t) =\omega_{R}(t)\omega_{M}(t)[ \mu_{R}(t)-\mu_{M}(t)]^2
\end{equation}

Where $\omega_{R}(t)$ and $\omega_{M}(t)$ are the probabilities of two classes separated by a threshold $t$ and $\mu_{R}(t)$ and $\mu_{M}(t)$ are the respective averages. All of them were calculated as follows:

\begin{equation}
    \begin{split}
        \omega_{R}(t)=\sum_{k=0}^{t-1}h_{i}(k)\\
        \omega_{M}(t)=\sum_{k=t}^{N_{bins}}h_{i}(k)\\
        \mu_{R}(t)=\dfrac{\sum_{k=0}^{t-1}k  \times h_{i}(k)}{\omega_{R}(t)}\\
        \mu_{M}(t)=\dfrac{\sum_{k=t}^{N_{bins}}k \times h_{i}(k)}{\omega_{M}(t)}
    \end{split}
\end{equation}
In order to speed up the processing time and reduce the number of points with low reflectivity values, we proposed an initial threshold similar to the one presented in \cite{huang2021} and \cite{huang2021b}. We calculated the mean value of reflectivity ($\bar{R}_{i}$) along with the variance ($VAR(R_{i})$) across all the points ($j$) in layer ($i$):
\begin{equation}
    \begin{split}
    \bar{R}_{i} = \frac{\sum_{j=1}^{N_{Pi}}R_{ij}}{N_{Pi}}\\
    VAR(R_{i}) = \frac{\sum_{j=1}^{N_{Pi}}R_{ij}^2}{N_{Pi}}-\bar{R}_{i}^2
    \end{split}
\end{equation}

Then, the computation of the adaptive threshold was executed as follows:
\begin{itemize}
    \item A layer $i$ was selected and the histogram $h_{i}(k)$ is calculated
    \item Set up an initial value for $\omega_{R,M}(0)$ and $\mu_{R,M}(0)$
    \item An iterative process was executed across all the possible values of $t$ starting with $t=\bar{R}_{i}+VAR(R_{i})$ as the initial value instead of 0. 
    \item $\sigma_{b}^{2}(t)$ was calculated. 
    \item The maximum value of $t$ was chosen as the final threshold.
    \item All the points with a reflectivity value equal or greater than the threshold were marked as road markings candidates.
\end{itemize}
Figure \ref{fig:alg_c} shows that most of the points that were not related to the road markings were removed from the original point cloud. However, there were still outliers  to be treated with the next step.

\subsection{Line fitting}
\label{sec:line}
We applied \gls{RANSAC} as an iterative method to find the line models which fit the road markings candidates resulting from the previous step. Once a line model was found, the supporting points were removed from the cloud and another line model was sought. The algorithm stopped when a maximum number of lines were found ($N_l$) or when the last line model found was supported by a small number of real points ($\leq N_p$). In Figure \ref{fig:alg_d}, the lines and their supporting points detected as road-markings are depicted in red. The lines and points that were rejected by the algorithm due to too few supporting points are depicted in light blue.

\section{Experiments}
\label{sec:experiments}
\subsection{Setup}
\label{subsec:setup}
\begin{figure}
	\centering
	\includegraphics[width=0.35\textwidth]{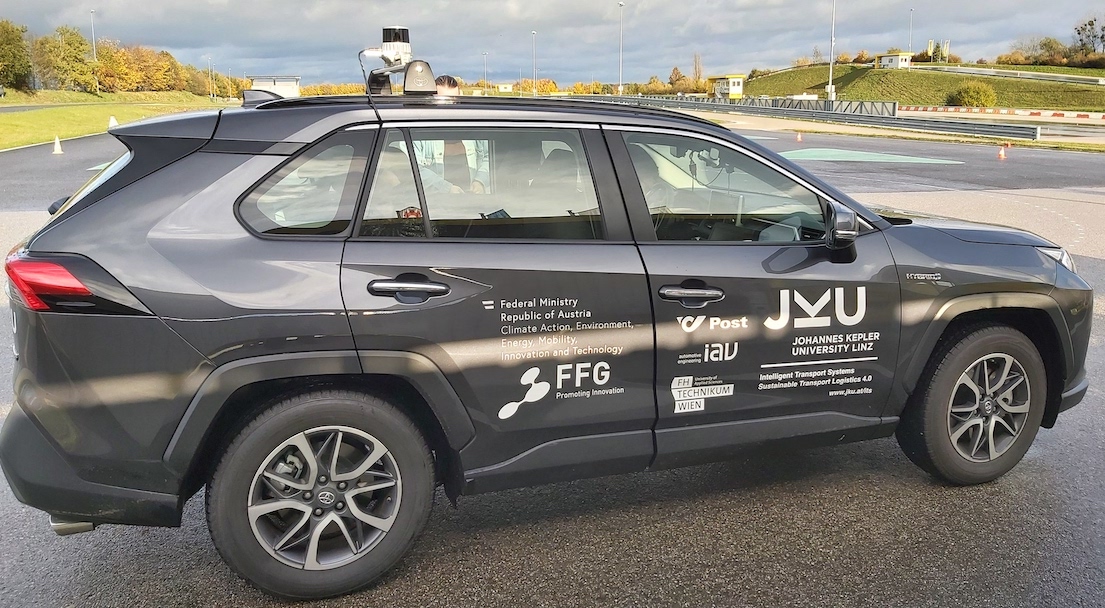}
	\caption{The JKU-ITS research vehicle \cite{Certad2022} was used to collect the data.}
	\label{fig:vehicle}
\end{figure}

To acquire the pertinent data to test the proposed approach we relied on the JKU-ITS research vehicle (See Figure \ref{fig:vehicle}). It consisted of a 2020 hybrid RAV4 from Toyota geared with an OS2-64  \gls{LIDAR} as well as an IMU, GPS, and a monocular camera \cite{Certad2022}. The acquisition system was based on \gls{ROS} running in a laptop connected to the sensors. The vehicle was used to record two different datasets on which the algorithm was tested afterwards. The final values set for the different variables described in section~\ref{sec:implementation} are depicted in Table \ref{tab:values}. 

\begin{table}[ht]
\caption{Values used during the experiments.}
\label{tab:values}
\begin{center}
\begin{tabular}{lc}
Variable    & Value     \\ \hline
 $Z_{l}$      & 1.44m          \\
 $Z_{u}$     & 2.44m          \\
 $TH_{plane}$  & 0.30m   \\
 $K_{RGC}$  & 30   \\
 $TH_{angle}$& $2^{\circ}$\\
 $TH_{curve}$& 1\\
 $N_{bins}$& 256\\
 $N_{L}$& 10 lines\\
 $N_{P}$& 10 points\\
 $TH_{lines}$  & 0.15m  
\end{tabular}
\end{center}
\end{table}

From each dataset, we randomly selected 200 frames and then ran them through the algorithm offline. The metrics we chose to evaluate the result were precision, recall, and F1-score which are widely used in the literature \cite{veronese2018,huang2021,huang2021b}.

To this end, we first defined: 
\begin{itemize}
    \item True Positives ($TP$): points marked as road-markings that supported a line that was in fact a lane-marking on the road.
    \item False Positive ($FP$) = points marked as road-markings that supported a line that was not a lane-marking.
    \item False Negatives ($FN$) = Points filtered out but that supported a line that was in fact a lane-marking on the road.
\end{itemize}

Then, we calculated the metrics according to \cite{veronese2018}:

\begin{equation}
   Precision = \frac{TP}{TP+FP}
\end{equation}

\begin{equation}
   Recall= \frac{TP}{TP+FN}
\end{equation}

\begin{equation}
   F1 score = 2\times\frac{Precision \times Recall}{Precision+Recall}
\end{equation}

In order to assess the differences between reflectivity and intensity data from the \gls{LIDAR}, we also ran the intensity data through our algorithm just by changing the data used to build the histogram described in section \ref{sec:thresholding}.

\subsection{Test track dataset}


This dataset was recorded in the Digitrans test track\cite{digitrans} located at St. Valentin, Austria, and the technical details are described below: 
\begin{itemize}
    \item 1100m total longitude (940m straight). 8m wide two-lanes track.
    \item There is a middle segment of 450m with 6-lanes and 20m wide.
    \item Left curve minimum radius 45m.
    \item Right curve (Roundabout) minimum radius 48m.
    \item Different types of markings: flat thin-layer and structured road markings, white and orange (Delivered and applied by SWARCO Road Marking Systems \cite{swarco}).
    \item Main material: asphalt. There are other side roads made out of concrete.
\end{itemize}

To validate our proposed algorithm we acquired data on lightning conditions that interfere with the camera of the car. We made sure to drive through the test track during sunset hours to obtain conditions where the sun faced directly to the front of the car when driving one way.

\begin{figure}[t]
	\centering
	\begin{subfigure}{0.237\textwidth}
		\framebox[\textwidth]{\includegraphics[width=\textwidth]{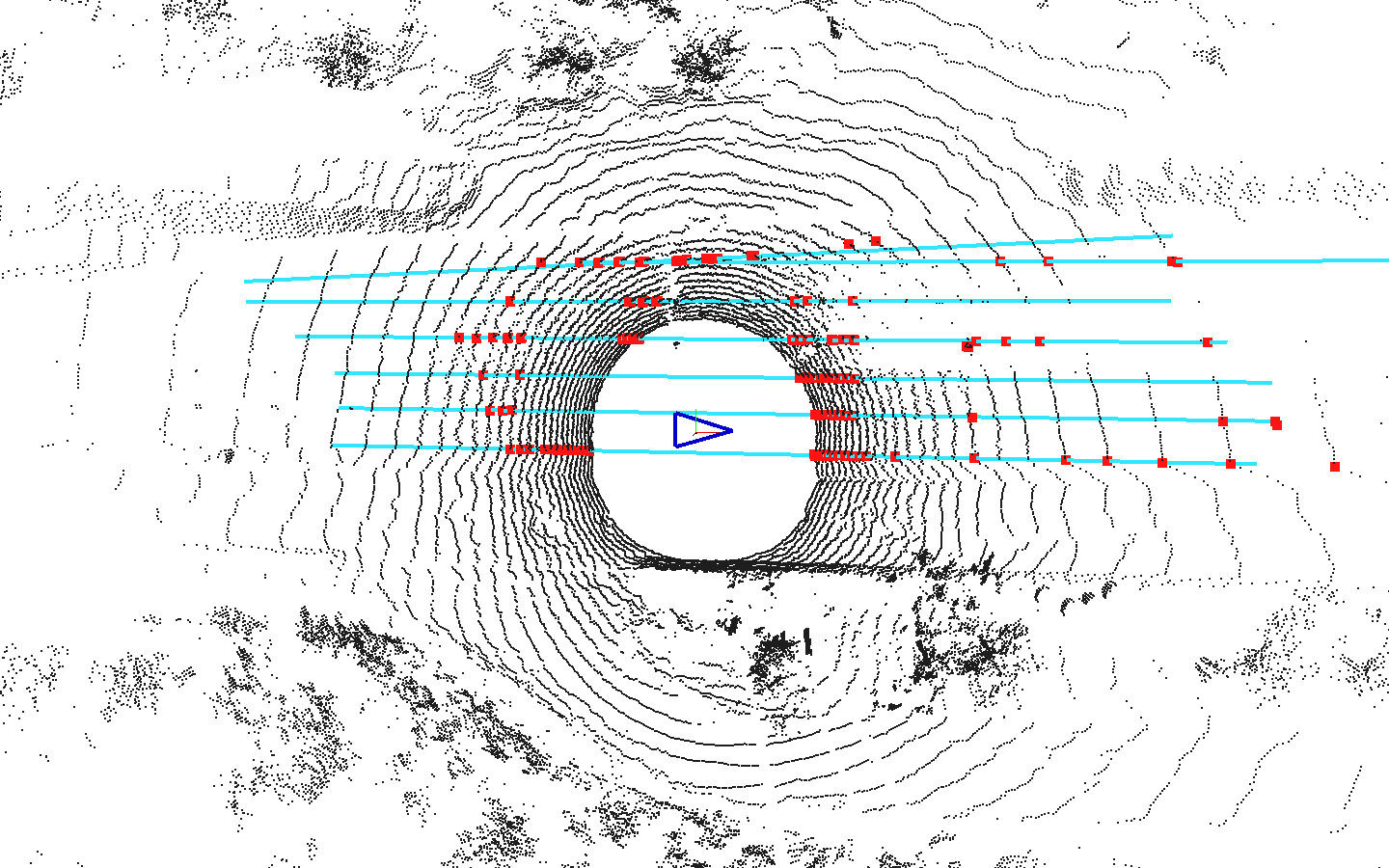}}
		\caption{}
	    \label{fig:pc_6_lane_1}
	\end{subfigure}
	\begin{subfigure}{0.237\textwidth}
		\framebox[\textwidth]{\includegraphics[width=\textwidth]{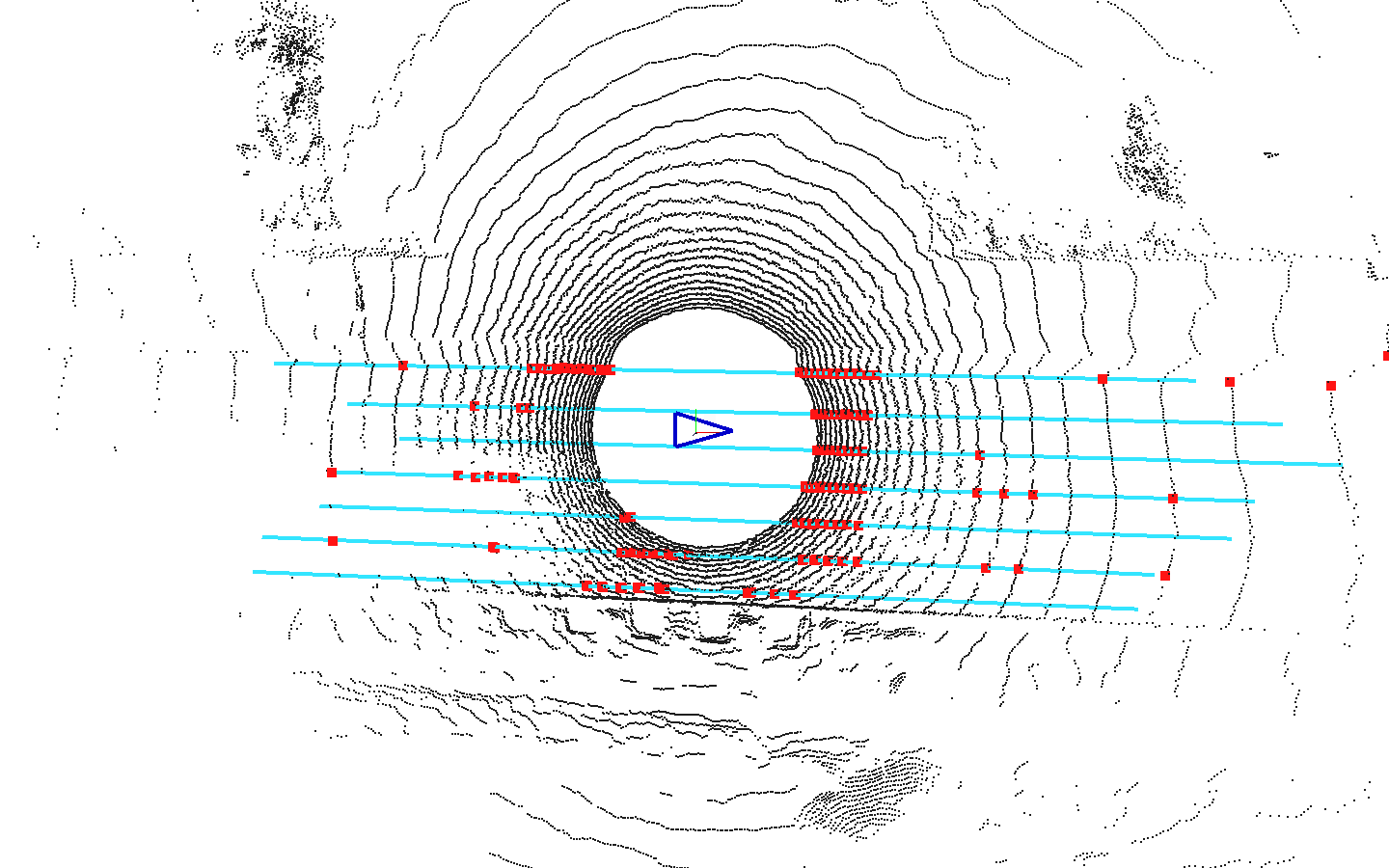}}
		\caption{}
	    \label{fig:pc_6_lane_2}
	\end{subfigure}
	\begin{subfigure}{0.237\textwidth}
		\includegraphics[width=\textwidth,trim={0 3cm 0 0},clip]{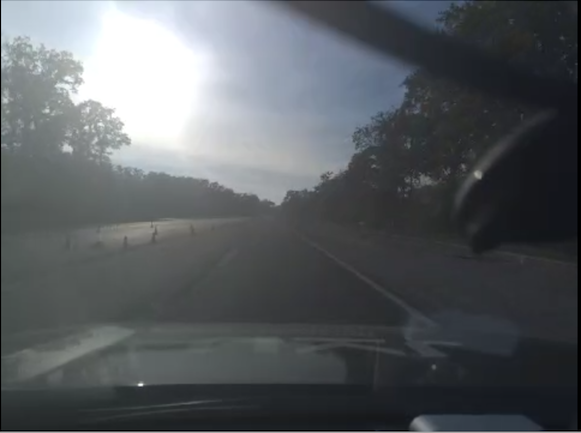}
		\caption{}
	    \label{fig:cam_6_lane_1}
	\end{subfigure}
    \begin{subfigure}{0.237\textwidth}
		\includegraphics[width=\textwidth,trim={0 3cm 0 0},clip]{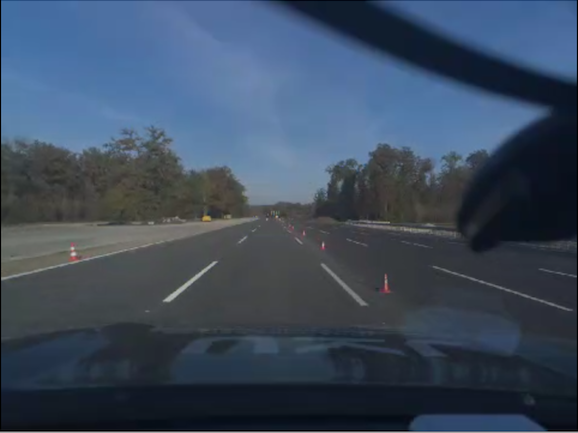}
		\caption{}
	    \label{fig:cam_6_lane_2}
	\end{subfigure}
	\caption{(a) and (b) depict the point clouds (dark gray) recorded when traversing the 6-lane segment of the test track. Red points are the detected road markings and blue lines indicate the line models. (c) and (d) shows the onboard camera images at the same time that (a) and (b) were recorded. In (c) the sun was in front of the vehicle making lane determination difficult with an image-only approach.}
	\label{fig:6_lanes}
\end{figure}
\subsection{Highway dataset}

This dataset was acquired in two different locations. The first one was a segment of the A7 highway around Linz in Austria during regular traffic conditions and typical daylight. The second location was a segment of the A3 highway near Bonn in Germany during regular traffic under low light level (cloudy evening). Unlike the first dataset, in this one there were occlusions caused by other vehicles on the road as well as road-markings in bad conditions. On the other hand, this dataset had no pronounced curvatures, which was advantageous for the representation of straight segments used in the developed algorithm.
\section{Results}
\label{sec:results}


\begin{table}[ht]
\caption{Results obtained across different datasets and lighting conditions}
\label{tab:results}
\begin{center}
\begin{tabular}{lcccc}
Dataset          & Channel        & Precision & Recall    & F1 score       \\ \hline
test track       & R   & 95.83\%   & 95.41\% & 95.62\%    \\ 
(high light level)         & I   & 89.69\%   & 93.88\% & 91.74\%      \\ \hline
test track       & R   & 97.90\%   & 94.48\% & 96.16\%    \\ 
(normal light level)    & I   & 96.88\%   & 91.94\% & 94.35\%      \\ \hline
highway - Linz   & R   & 98.97\%   & 89.03\% & 93.74\%      \\
(normal light level)   & I   & 89.87\%   & 89.06\% & 89.46\%     \\ \hline
highway - Bonn   & R   & 96.68\%   & 95.14\% & 95.90\%      \\
(low light level)      & I   & 91.69\%   & 92.20\% & 91.94\%     \\ \hline
\end{tabular}
\end{center}
\end{table}

\begin{table}[ht]
\caption{Overall results in comparison to other works}
\label{tab:results2}
\begin{center}
\begin{tabular}{lcccc}
Method          & Channel        & Precision & Recall    & F1 score       \\ \hline 
Our method (overall) & R   & 97.04\%   & 94.03\% & 95.51\%      \\
                 & I   & 91.67\%   & 91.82\% & 91.74\%       \\ \hline           
Method presented& I   & 90.92\%   & 92.84\% & 91.82\%      \\
  in\cite{huang2021}.       & I   & 95.21\%   & 84.66\% & 89.49\%                     
\end{tabular}
\end{center}
\end{table}

\begin{figure}[ht]
	\centering
	\begin{subfigure}{0.237\textwidth}
		\framebox[\textwidth]{\includegraphics[width=\textwidth]{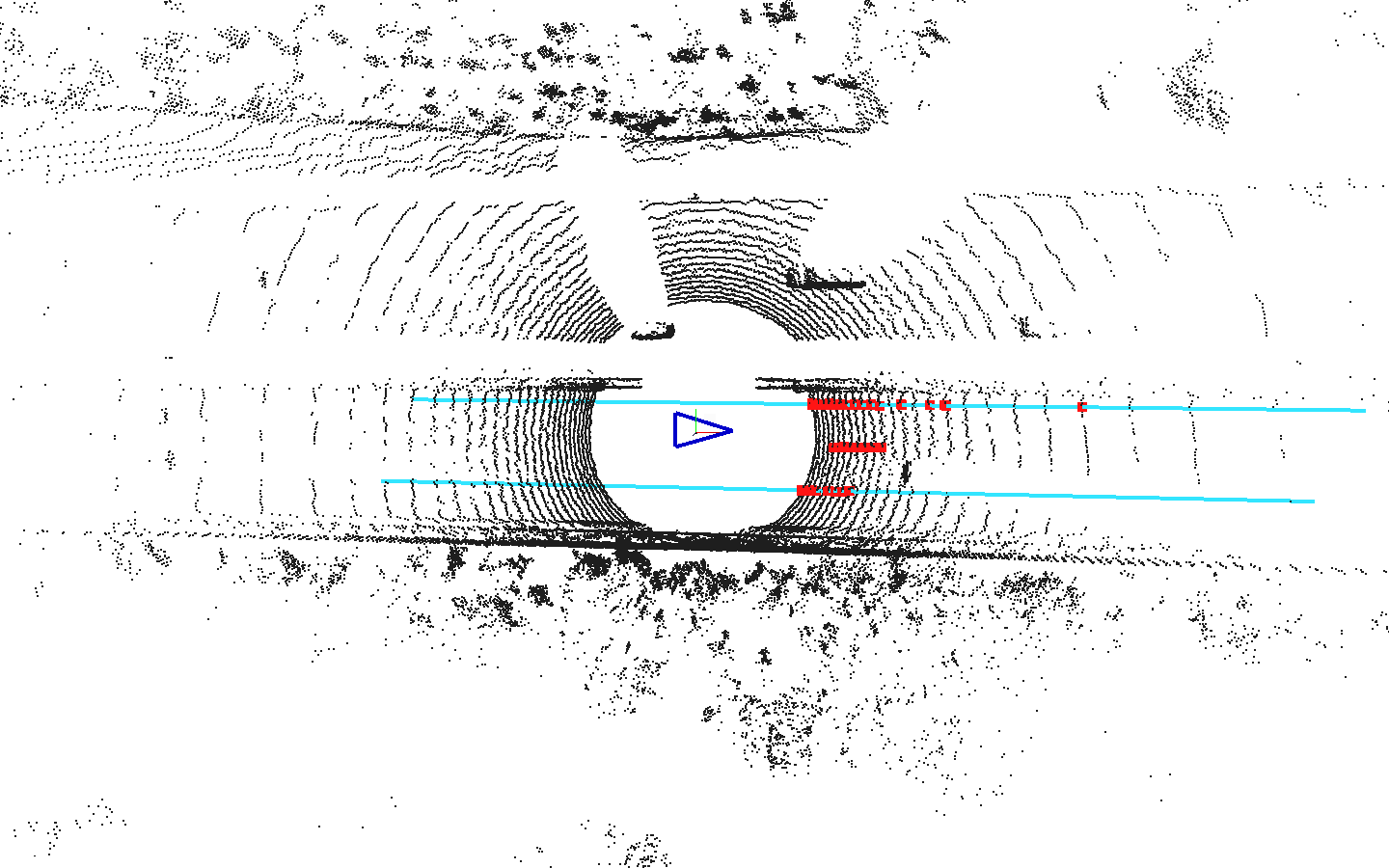}}
		\caption{}
	    \label{fig:high_2}
	\end{subfigure}
	\begin{subfigure}{0.237\textwidth}
		\framebox[\textwidth]{\includegraphics[width=\textwidth]{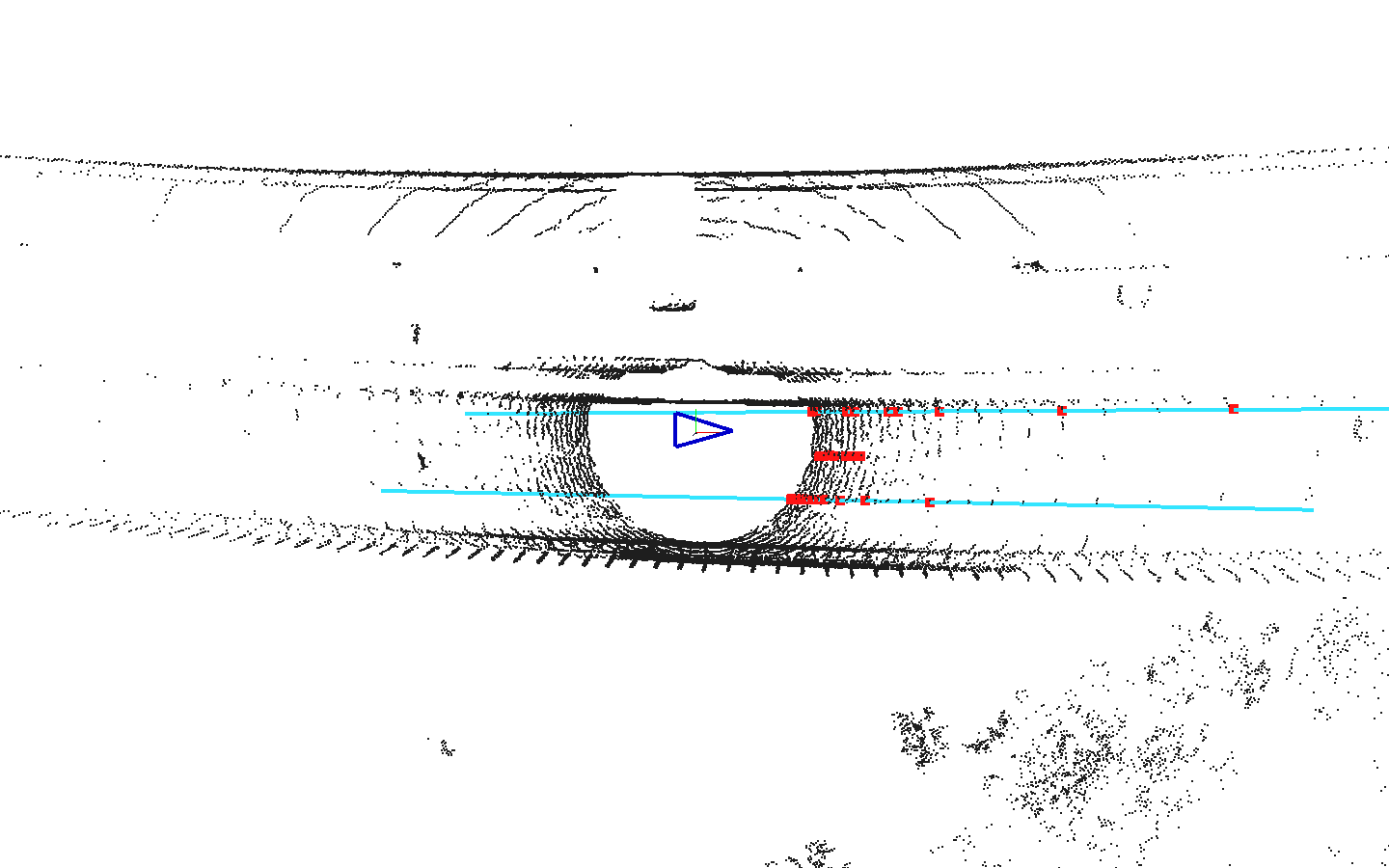}}
		\caption{}
	    \label{fig:high_3}
	\end{subfigure}
	\caption{Two point clouds (dark gray) recorded when traversing the highway at high speed. Red points are the detected road markings and the blue lines indicate the estimated lane-markings.}
	\label{fig:highway}
\end{figure}
The detailed results from both datasets are depicted in Table \ref{tab:results}. The best results were obtained on the test track where road markings are kept in good condition. In the highway dataset, there is a slight reduction in the F1 score due to two reasons. First, the road markings conditions were not the same as in the test track. Second, the vehicle was driven at high speed. This made the detection of the points over the dashed center-line difficult. At high speed, the points supporting the dashed center-lines are too sparse and the algorithm filtered them out (as can be seen in Figure \ref{fig:highway}). As can be seen in Figure \ref{fig:6_lanes}, our algorithm was not restricted to identifying the lane-lines where the vehicle was traversing but almost all the supporting points of the available lane-lines on the road. It is also reliable under different lighting conditions, with no detectable differences along the three lighting levels that were tested: high light level when the vehicle was driven with the sun in front; normal lighting level (daylight); and low level of light (cloudy evening). The results depicted in Table \ref{tab:results} and Table \ref{tab:results2} show a slight advantage when using reflectivity data instead of intensity data when both channels are run through the exact same algorithm thus confirming the hypothesis proposed in this work.

The overall results of our method are depicted in Table \ref{tab:results2} along with a comparison with the results obtained in \cite{huang2021}. Even though our method exhibits slightly better results, both methods were tested in different datasets thus a point-to-point comparison is not possible.

An attempt was made to test the algorithm in the rain, however, the \gls{LIDAR} sensor used does not have the ability to detect the road when it is covered with a layer of water.
\section{Conclusion and Future Work}
\label{sec:conclusion}
In this paper, we introduced a method to detect road-marking points from \gls{LIDAR} data. In contrast to currently available methods, our procedure was not limited to the current lane the vehicle is traversing and was able to extract the road markings from all the lanes of the road which is an important feature for lane changing algorithms. The results showed an improvement just by using the reflectivity data directly provided by the \gls{LIDAR} instead of the raw intensity data. 

Currently, we are working to add a tracking system to preserve the road-marking information over time and reduce the detected false negatives. In the near future, we plan to substitute the line models described in section \ref{sec:line} with semicircular arcs to improve the detection in curves. We also plan to test the procedure in other environmental conditions like haze, light snow, and nighttime.

\section*{ACKNOWLEDGMENT}

This work was partially supported by the Austrian Ministry for Climate Action, Environment, Energy, Mobility, Innovation and Technology (BMK) Endowed Professorship for Sustainable Transport Logistics 4.0., IAV France S.A.S.U., IAV GmbH, Austrian Post AG and the UAS Technikum Wien. It was additionally supported by the Austrian Science Fund (FWF), project number P 34485-N.

\bibliographystyle{IEEEtran}
\bibliography{paper}
\end{document}